\def\year{2020}\relax
\documentclass[letterpaper]{article} 
\usepackage{aaai20}  
\usepackage{times}  
\usepackage{helvet} 
\usepackage{courier}  
\usepackage[hyphens]{url}  
\usepackage{graphicx} 
\usepackage{graphicx}  
\newcommand\figref{Figure~\ref}
\usepackage{bm}
\usepackage{algorithm}
\usepackage{algorithmic}
\usepackage{booktabs} 
 \usepackage{amssymb}
\usepackage{lipsum}
\usepackage{amsmath}
\usepackage{amssymb}
\usepackage{mathrsfs}
\usepackage{subfigure}
\usepackage{array}

\DeclareMathOperator*{\argmin}{arg\,min}
\title{Self-Adaptive Label Augmentation for Semi-supervised Few-shot Classification}
\author{Xueliang Wang, Jianyu Cai, Shuiwang Ji\textsuperscript{\dag}, Houqiang Li, Feng Wu, Jie Wang\\
{University of Science and Technology of China, Texas A\&M University\textsuperscript{\dag}}\\
\{xlwang95, caijianyu1997\}@mail.ustc.edu.cn, sji@tamu.edu\\
\{lihq,  fengwu, jiewangx\}@ustc.edu.cn}

\begin{document}

\maketitle

\begin{abstract}
Few-shot classification aims to learn a model that can generalize well to new tasks when only a few labeled samples are available. To make use of unlabeled data that are more abundantly available in real applications, Ren et al. \shortcite{ren2018meta} propose a semi-supervised few-shot classification method that assigns an appropriate label to each unlabeled sample by a manually defined metric. However, the manually defined metric fails to capture the intrinsic property in data. In this paper, we propose a \textbf{S}elf-\textbf{A}daptive \textbf{L}abel \textbf{A}ugmentation approach, called \textbf{SALA}, for semi-supervised few-shot classification. A major novelty of SALA is the task-adaptive metric, which can learn the metric adaptively for different tasks in an end-to-end fashion. Another appealing feature of SALA is a progressive neighbor selection strategy, which selects unlabeled data with high confidence progressively through the training phase. Experiments demonstrate that SALA outperforms several state-of-the-art methods for semi-supervised few-shot classification on benchmark datasets.
\end{abstract}

\section{Introduction}\label{submission}

Deep learning methods have achieved great success in a wide range of
applications, including visual recognition
\cite{donahue2014decaf,he2016deep}, machine translation
\cite{cho2014learning,sutskever2014sequence},
speech recognition \cite{hinton2012deep,dahl2012context}, etc.
However, these models need large amounts of high-quality labeled
data and many iterations to train the parameters. This requirement prevents the
potential applications of the existing deep learning methods to the few-shot
learning regime, in which only a few labeled examples are available
for each category. However, humans
can learn new skills and concepts quickly from a small number of
labeled samples. For example, children can recognize apples after
being presented with a few instances of it. We wonder whether the
machines can acquire the ability to learn efficiently from only a
few labeled samples.

Traditional work
cannot handle few-shot learning problems effectively. As claimed
by Vinyals et al. \shortcite{vinyals2016matching}, the traditional
\emph{fine-tuning} models in deep learning suffer the problem of
\emph{over-fitting}. To tackle this problem, there has been a significant body of research work on
few-shot learning
\cite{lake2011one,finn2017model,garcia2017few,sung2018learning,oreshkin2018tadam,bertinetto2018meta,ye2018learning},
which aims to learn a model that can generalize well to new classes
with only a few labeled samples. MAML \cite{finn2017model}
aims to find more transferable initial parameters. Vinyals et al. \shortcite{vinyals2016matching} and  Snell et al. \shortcite{snell2017prototypical}
propose to learn a shared metric, which can be seen as a
nearest-neighbor classifier. Moreover,
Santoro et al. \shortcite{santoro2016meta} and Mishra et al. \shortcite{mishra2017meta} propose to train a generic
inference network. These approaches have achieved great progress in solving few-shot learning
problems.

While promising, many existing methods are limited to supervised learning settings, as they require labeled data for training. However, unlabeled data are more abundantly available in real-world applications and can be useful to improve the performance. This naturally motives the idea of introducing the semi-supervised learning techniques to facilitate few-shot learning. Recently, Ren et al. \shortcite{ren2018meta} propose a semi-supervised few-shot classification method based on the Prototypical
Network \cite{snell2017prototypical}, which learns the prototype representation of each class by both labeled and unlabeled data. Zhang et al. \shortcite{zhang2018metagan} introduce new loss terms to leverage unlabeled instances based on traditional GAN \cite{goodfellow2014generative}. Liu et al. \shortcite{liu2019fewTPN} propose a Transductive Propagation Network to propagate labels from seen labeled instances to unseen unlabeled instances.



However, most existing methods use a manually defined metric (e.g., the Euclidean distance) to assign a label, i.e., the so-called \emph{pseudo label}, to each unlabeled sample, thereby they fail to take into account the intrinsic property in data. This may lead to an inappropriate labeling that hurts the classification performance. Moreover, through all episodes, existing methods use all of the sampled pseudo-labeled data---including the data with low confidence pseudo labels---to train the parameters. Consequently, the data with unreliable pseudo labels may seriously mislead the model, especially during the early stage of training.

To tackle the aforementioned challenges, we propose a \textbf{S}elf-\textbf{A}daptive \textbf{L}abel \textbf{A}ugmentation
approach, called \textbf{SALA}, for semi-supervised few-shot classification. A major novelty of SALA is that for each learning task, our approach can effectively learn the task-adaptive metric in an end-to-end fashion by incorporating the intrinsic data property. Moreover, inspired by the observation that the pseudo-labels will be more reliable as the training course proceeds, we further propose a novel progressive neighbor selection strategy; that is, SALA only selects a small number of samples with the most reliable pseudo labels for training and increases the number of them progressively.  Experiments on benchmark datasets  \cite{liu2019fewTPN,ren2018meta,zhang2018metagan} demonstrate that SALA
significantly outperforms several state-of-the-art methods.


We organize the rest of this paper as follows. We first review the related work in Section \ref{related work}. Then, we describe the problem and introduce the proposed SALA in Section \ref{Semi-supervised Few-shot Learning}. We show experiments in Section \ref{Experiments} and finally conclude this paper in Section \ref{Conclusion}.


\section{Related work}\label{related work}






 The few-shot learning methods fall into roughly three categories \cite{chen2019closer,rusu2018meta,wang2019generalizing}, which are optimization-based methods, memory-based methods, and metric-based methods.


 
  The optimization-based methods tackle the few-shot learning problem by \emph{learning to fine-tune} \cite{chen2019closer}. One approach tries to find transferable and sensitive initial parameters, which can quickly adapt to new tasks through a few gradient-descent steps \cite{maclaurin2015gradient,finn2017model,qiao2018few,nichol2018reptile,rusu2018meta}. Another line of research is to train an \emph{optimizer}---instead of hand-designed update rules (e.g., stochastic gradient decent)---for efficient parameter updates with only a few samples \cite{DBLP:journals/corr/AndrychowiczDGH16,ravi2016optimization,wang2019tafe}. However, the aforementioned approaches suffer from the need to fine-tune on the target domain \cite{sung2018learning} and have difficulties in transferring to the new domain with new classes \cite{chen2019closer}.  

  The memory-based methods deal with the few-shot learning problem by \emph{learning to remember}. These approaches try to make use of the knowledge from previous tasks such that the learned models can adapt to new tasks quickly. Specifically, these methods accumulate knowledge---e.g., important training samples or weight-update mechanism---from prior tasks by memory-augmented neural networks \cite{munkhdalai2017meta,santoro2016meta}.  While promising, these approaches confront the challenge that it is difficult to reliably store all the important and historical information of relevance without forgetting \cite{sung2018learning}. 
  
  
  The metric-based methods address the few-shot learning problem by \emph{learning to compare} \cite{chen2019closer}. The motivation is that samples from the same class should be close to one another and far away from the samples from other classes. Koch et al. \shortcite{koch2015siamese} propose to determine if samples are from the same class by Siamese Neural Networks. Vinyals et al. \shortcite{vinyals2016matching} and Snell et al. \shortcite{snell2017prototypical} use cosine similarity and Euclidean distance, respectively, to compute the similarity between samples. Sung et al. \shortcite{sung2018learning} propose to learn a deep distance metric by the relation network to compare the similarity between samples. 
    

  
The aforementioned approaches are all under the supervised learning settings, which require labeled data for training. To make use of unlabeled data, which is more common in many real applications, extensive research efforts have been devoted to introducing the semi-supervised techniques to the few-shot learning problems. Ren et al. \shortcite{ren2018meta} extend the Prototypical Network to semi-supervised few-shot classification, which learns the prototype representations based on both labeled and unlabeled data. Zhang et al. \shortcite{zhang2018metagan} introduce new loss terms to leverage unlabeled instances by the traditional GAN \cite{goodfellow2014generative}. Liu et al. \shortcite{liu2019fewTPN} propose the Transductive Propagation Network that learns a graph construction module to propagate labels from labeled instances to unlabeled instances.

 \begin{figure*}[ht]
     \centering
     \includegraphics[scale=0.33]{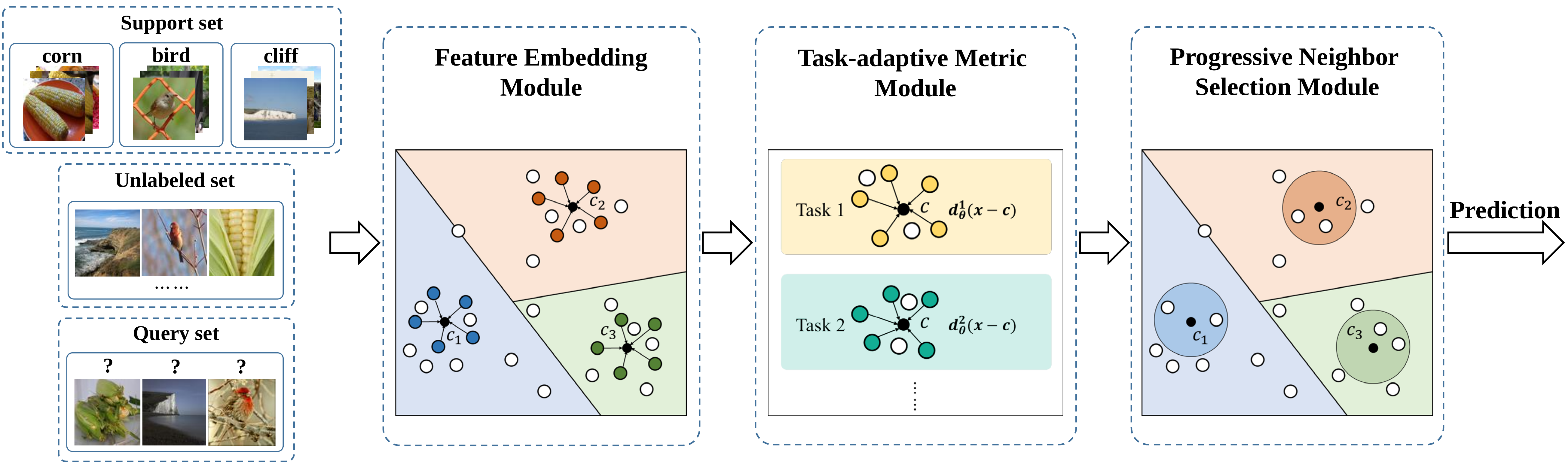}
     \caption{The framework of SALA. SALA mainly includes three modules: the feature embedding module, the task-adaptive metric module, and the progressive neighbor selection module. The colored points and white points denote the feature vectors of the labeled and unlabeled samples, respectively.}
     \label{fig:figure3}
 \end{figure*}

\section{Self-adaptive Label Augmentation}\label{Semi-supervised Few-shot Learning}
In this section, we first introduce the basic settings of semi-supervised few-shot classification following Ren et al. \shortcite{ren2018meta} and Liu et al. \shortcite{liu2019fewTPN}. Then, we present the proposed SALA in details.



\subsection{Problem Settings}\label{Problem Definition}
Few-shot classification aims to train a model on a large dataset $D_{train}$ that can generalize well to a new dataset $D_{test}$ with \emph{another set of classes} \cite{ren2018meta}. Semi-supervised few-shot classification implies that only a few samples have labels in these datasets. In this paper, we consider the $N$-way $K$-shot episodes and follow the two settings considered in the seminal paper by Ren et al. \shortcite{ren2018meta}. 

The first one assumes that, in each episode, all the unlabeled samples come from the same set of classes as the labeled samples.
Specifically, in each episode, we first randomly choose $N$ classes from $D_{train}$, from each of which we then randomly select $K$ labeled samples to construct the support set $\mathcal{S}=\{(\bm{x}_1,y_1),(\bm{x}_2,y_2),...,(\bm{x}_{N\times K},y_{N \times K})\}$. Then, we select $Q$ different samples in the same $N$ classes to construct the query (test) set $\mathcal{Q}=\{(\bm{x}_1^*,y_1^*),(\bm{x}_2^*,y_2^*),...,(\bm{x}_Q^*,y_Q^*)\}$ for evaluation. We further construct another unlabeled set  $\mathcal{U}=\{\widetilde{\bm{x}}_1,\widetilde{\bm{x}}_2,...,\widetilde{\bm{x}}_M\}$ in each episode, where the unlabeled samples in $\mathcal{U}$ come from the same $N$ classes as the labeled samples in $\mathcal{S}$. 


The second one is more challenging, as the unlabeled samples in $\mathcal{U}$ may come from classes unseen in $\mathcal{S}$ and thus be called \emph{distractors}. In this case, we generate the support set $\mathcal{S}$ and the query set $\mathcal{Q}$ in the same way as the setting mentioned above. However, we need to take extra care of those distractors.

In each training episode, we feed ($\mathcal{S}$, $\mathcal{U}$) to the model and update its parameters to minimize the prediction loss on $\mathcal{Q}$. After the training phase, we evaluate the trained model on $D_{test}$ in the similar $N$-way $K$-shot paradigm. Notice that, in subsequent sections, we use the single word \emph{task} to refer to the \emph{$N$-way $K$-shot learning task} in each episode \cite{vinyals2016matching,finn2017model,snell2017prototypical} for convenience.

\subsection{The Framework of SALA}\label{structure}

As shown by Figure \ref{fig:figure3}, SALA mainly consists of three modules: the feature embedding module, the task-adaptive metric module, and the progressive neighbor selection module. Recall that, the major novelties of the proposed SALA lie in the latter two modules.

\textbf{The feature embedding module} aims to find appropriate feature vectors for input images and then computes the prototype representation for each class.


\textbf{The task-adaptive metric module} aims to learn a task-adaptive metric rather than the existing manually defined metric to better capture the intrinsic property in each $N$-way $K$-shot learning task.

\textbf{The progressive neighbor selection module} aims to reliably update the prototype representations; that is, it selects only a small number of unlabeled samples with high confidence at the early stage of the training phase and increases the number of them progressively as the training proceeds. 

In each episode, SALA first finds the prototype representations (cluster centers) of the support set by the feature embedding module. Then, based on the task-adaptive metric---which adapts to different tasks automatically---SALA assigns \emph{pseudo labels} to the unlabeled samples and applies the progressive neighbor selection module to reliably update the prototype representations. We can then predict the labels on the query set with the refined prototype representations. We detail the above three modules in Sections \ref{Feature Embedding Module}, \ref{Task-adaptive Metric Module}, and \ref{Pseudo label propagation module}, respectively. Finally, we introduce the loss function and summarize SALA in Section \ref{loss function}.

\subsection{Feature Embedding Module}\label{Feature Embedding Module}
To compare with the existing methods fairly, we use ConvNet \cite{snell2017prototypical,ren2018meta,liu2019fewTPN} as the feature extractor, which includes four-layer convolutional blocks. Each block consists of a 64-filter $3\times 3$ convolution, a batch normalization layer, a ReLU layer, and a $2\times 2$ max-pooling layer.

In each episode, let $(\bm{x}_{s},y_s)\in \mathcal{S}$, $(\bm{x}_q^*,y_q^*)\in \mathcal{Q}$ and $\widetilde{\bm{x}}_u\in \mathcal{U}$. The feature embedding module finds the feature vector $\bm{g}_\theta(\bm{x})\in \mathbb{R}^d$ of each sample, where $\theta$ denotes the parameters of the feature embedding module and $d$ is the dimension of the feature vector. Then, we compute the prototype representations of the labeled samples \cite{snell2017prototypical}
\begin{align}\label{cluster}
    \bm{c}_i = \frac{\sum_{s} \bm{g}_{\theta}(\bm{x}_s) z_{s,i}}{\sum_s z_{s,i}},\,\,z_{s,i} = \mathbb{I}[y_s = i], 
\end{align}
where $i=1,2,\ldots,N$ and
$\mathbb{I}[\cdot]$ is the indicator function:
\begin{align}
\mathbb{I}[y_s=i]=
\begin{cases}
1,\,y_s=i,\\
0,\,y_s\neq i.
\end{cases}
\end{align}

\subsection{Task-adaptive Metric Module}\label{Task-adaptive Metric Module}


As mentioned before, Ren et al. \shortcite{ren2018meta} propose a manually defined metric to measure the similarity between samples and the prototype representations:
\begin{align} \label{euclidean_distance}
        {\rho}_{\theta}(\bm{x}, \bm{c}_i) = \| \bm{g}_\theta(\bm{x}) - \bm{c}_i \|_2^2.
\end{align}

\eqref{euclidean_distance} implies that different features are equally important in measuring the similarity. However, in many applications, empirical observations frequently show that some features are more important than others \cite{DataMiningAggarwal}.  Inspired by this observation and further notice that the intrinsic property in different tasks may vary significantly, we propose the novel task-adaptive metric as follows:
    \begin{align}\label{adaptive distance}
        d_{\theta}(\bm{x}, \bm{c}_i) = \sum_{j=1}^{d} \alpha^{T}_{ij}  ([\bm{g}_\theta(\bm{x})]_j - [\bm{c}_i]_j)^2,
    \end{align}
where $\alpha^{T}_{ij}$ is the positive weight to be learned in each task. The superscript of $\alpha^{T}_{ij}$ emphasizes that this weighting factor can be different across different tasks.

The weighting factors in \eqref{adaptive distance} play a fundamentally important role in the proposed task-adaptive metric. Intuitively, the major novelty of the proposed task-adaptive metric is that, it can effectively capture the intrinsic property---i.e., the common meta-knowledge \cite{vilalta2002perspective}---across different tasks. This is achieved by generating a set of weighting factors $\alpha^{T}_{ij}$ that can adapt to various tasks according to their intrinsic nature. Therefore, through the task-adaptive weighting factors, the proposed task-adaptive metric naturally encodes feature importance for different classes and different tasks. 

\begin{figure*}[]
    \centering{
                \subfigure[]{ \label{task-adaptive}
            \includegraphics[height=0.66\columnwidth]{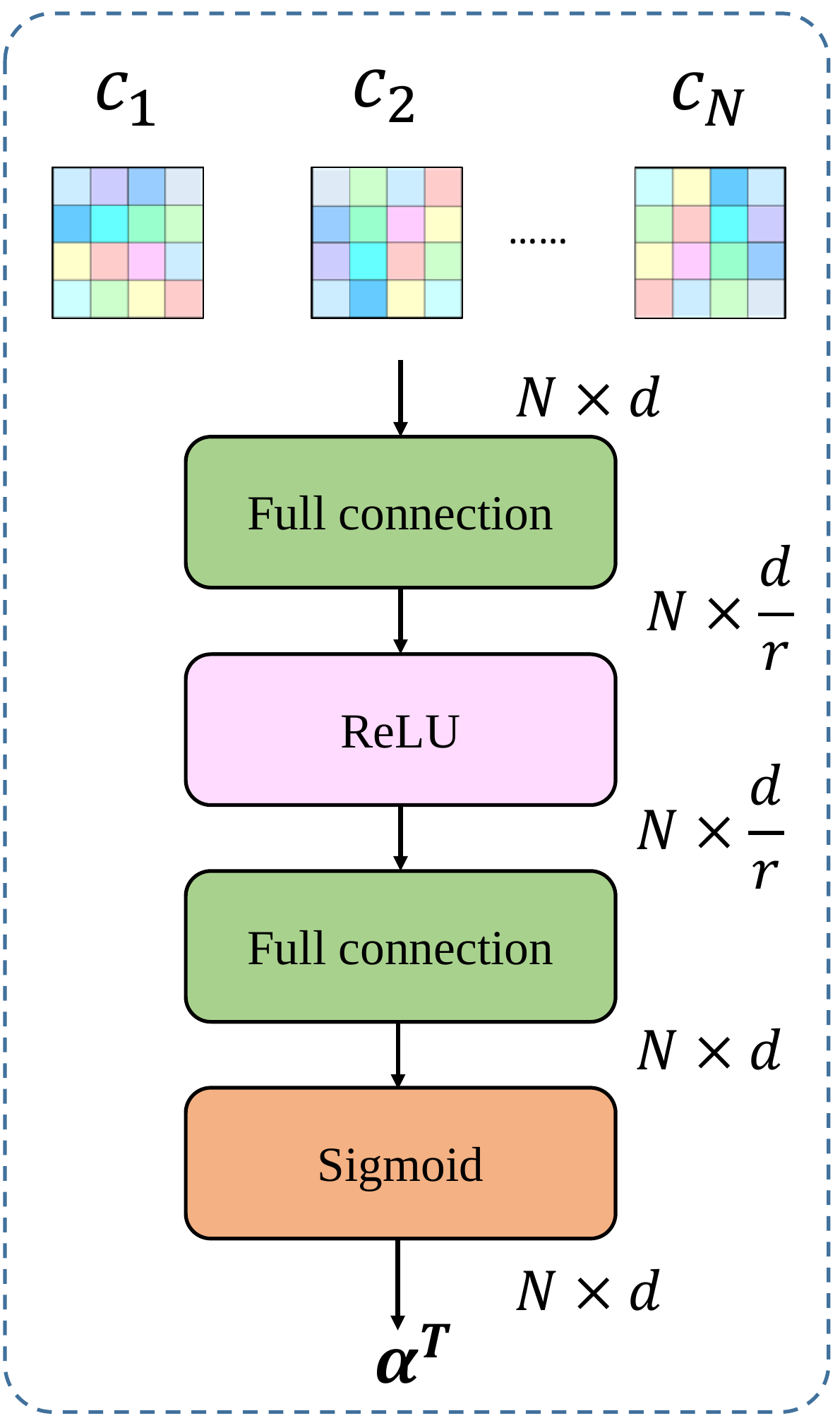}
        }\hspace{34mm}
        \subfigure[]{ \label{details}
            \includegraphics[height=0.66\columnwidth]{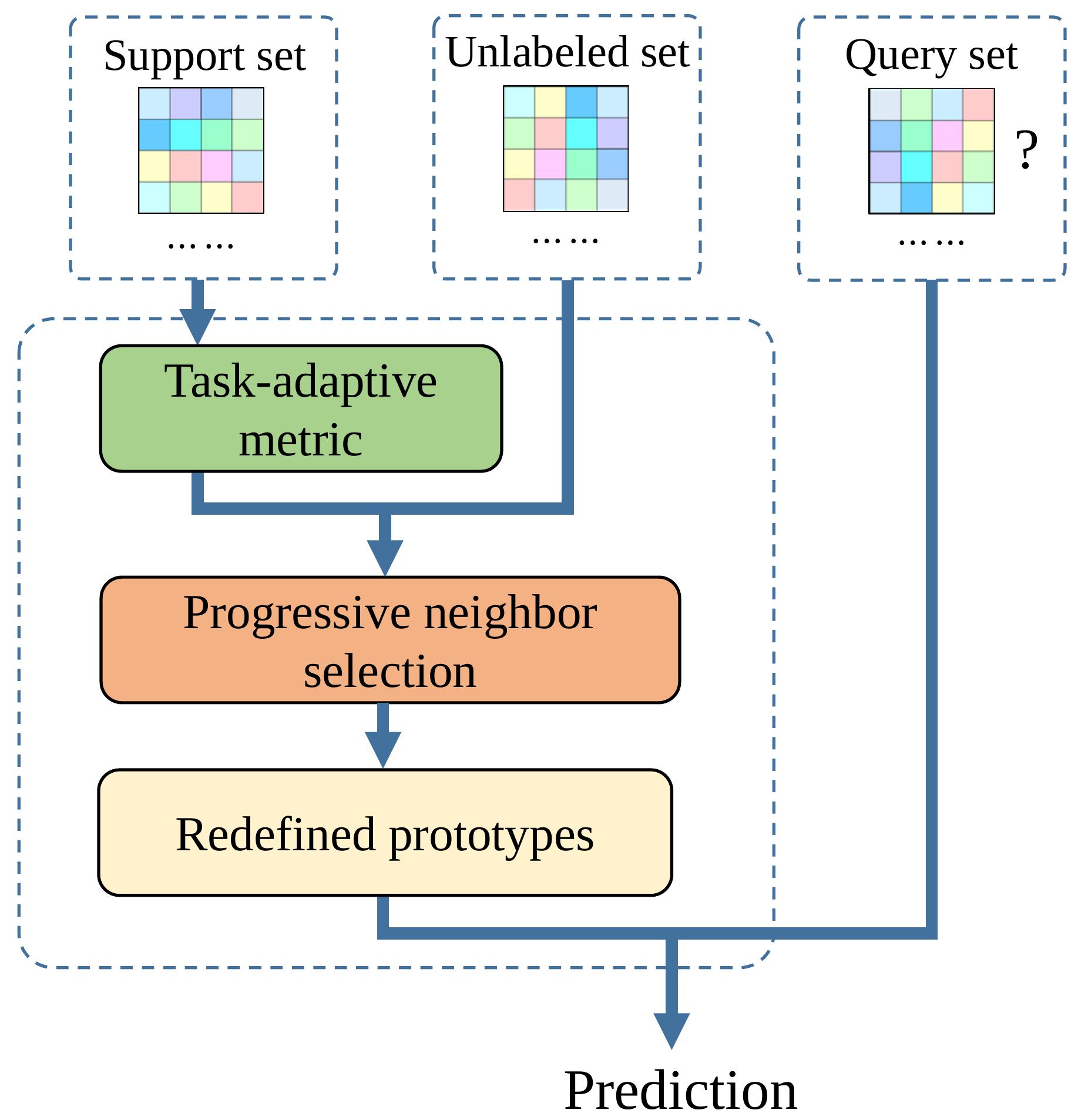}
        }
    }
    \caption{We show the structure of the SE network in Figure \ref{task-adaptive}. Figure \ref{details} illustrates how the task-adaptive metric module and the progressive neighbor selection module interact with each other.}
    \label{}
\end{figure*}

Specifically, in each episode, we feed the $N$ prototype representations into a simple four-layer Squeeze-and-Excitation (SE) network \cite{hu2018senet} to generate the weighting factors $\alpha^{T}_{ij}$. Figure \ref{task-adaptive} illustrates the structure of the SE network, which consists of a dimensionality-reduction layer with reduction ratio $r$, a ReLU layer, a dimensionality-increasing layer returning to the original dimension, and a sigmoid layer.

\subsection{Progressive Neighbor Selection Module}\label{Pseudo label propagation module}
In order to use the unlabeled samples to facilitate the few-shot classification problems, commonly used semi-supervised few-shot learning approaches \cite{ren2018meta} generate pseudo-labels to unlabeled samples based on their distances to the prototype representations. Thus, it is critical to keep the prototype representations at high levels of confidence such that the pseudo-labels may correctly indicate the true nature of the unlabeled samples. The progressive neighbor selection module aims to reliably update the prototype representations and further refine them progressively.




Specifically, for a given task, once we get the prototype representations of the labeled samples by \eqref{cluster}, we can compute the similarity $d_{\theta}(\widetilde{\bm{x}}_u, \bm{c}_i)$ between the unlabeled sample $\widetilde{\bm{x}}_u$ and $\bm{c}_i$ by \eqref{adaptive distance}. Then, we define the probability that $\widetilde{\bm{x}}_u$ belongs to the $i^{th}$ class by
\begin{align}\label{compute_probability}
    p_{i}(\widetilde{\bm{x}}_u) = \frac{\exp(-d_{\theta}(\widetilde{\bm{x}}_u, \bm{c}_i))}{\sum_{i=1}^{N} \exp(-d_{\theta}(\widetilde{\bm{x}}_u, \bm{c}_i))},\,i=1,2,\ldots,N.
\end{align}

Existing methods such as \cite{ren2018meta} use the average---weighted by the probabilities computed by \eqref{compute_probability}---of all the unlabeled samples $\widetilde{\bm{x}}_u$ in $\mathcal{U}$ to update the prototype representation $\bm{c}_i$. However, during the early stage of the training, the model has not been well trained, which may lead to low confidence level of the similarity measured by the metric. This can in turn seriously mislead the model in the subsequent training stage. 

To address this challenge, we propose a novel progressive neighbor selection strategy to reliably update the prototype representations; that is, we only use the unlabeled samples with high confidence to update the prototype representations. Specifically, for each $\widetilde{\bm{x}}_u\in \mathcal{U}$, we define 
\begin{align}\label{confidence}
    &d^*_{\theta}(\widetilde{\bm{x}}_u)=\min\{d_{\theta}(\widetilde{\bm{x}}_u, \bm{c}_i):i=1,2,...,N\}
\end{align}
\begin{align}\label{confidentclass}
    &\pi^*_{\theta}(\widetilde{\bm{x}}_u)=\argmin\{d_{\theta}(\widetilde{\bm{x}}_u, \bm{c}_i):i=1,2,...,N\}
\end{align}
Intuitively, a small value of $d^*_{\theta}(\widetilde{\bm{x}}_u)$ implies that the unlabeled sample $\widetilde{\bm{x}}_u$ is close to one of the prototypes, say $\bm{c}_k$, where $k=\pi^*_{\theta}(\widetilde{\bm{x}}_u)$. In other words, it is of high confidence level that the unlabeled sample $\widetilde{\bm{x}}_u$ comes from the $k^{th}$ class. Moreover, we note that, the prediction performance of the model keeps improving as the training proceeds, and so does the confidence level of the similarity measured by the metric. Therefore, we can increase the number of unlabeled samples to update the prototype representations progressively. This leads to the proposed progressive neighbor selection strategy as follows.


Let $M_0$ be the largest number of unlabeled samples we select from $\mathcal{U}$ and $w(t)$ \cite{laine2016temporal} be given by
\begin{align}
    w(t)=\exp[-\eta\cdot(1-t)^2],
\end{align}
where $t\in[0,1]$ is a linear function of the index of episodes, and $\eta$ is the parameter that controls the selection rate.
In each episode, we select the top $n$ unlabeled samples with the highest confidence level measured by \eqref{confidence} from $\mathcal{U}$, where $n$ is given by
\begin{align}
    n = \min\{w(t)*M_{0},M_{0}\}.
\end{align}
Denote the set of selected unlabeled samples by $\mathcal{U}'$. 
We update the propotype representations $\bm{c}_i$, $i=1,2,...,N$, by
\begin{align}\label{update prototype representation}
    \bm{c}_i\leftarrow\frac{\sum_{s}  z_{s,i} \bm{g}_{\theta}(\bm{x}_s)+\sum_{u\in \mathcal{U}'} p_{i}(\widetilde{\bm{x}}_u) \bm{g}_\theta(\widetilde{\bm{x}}_u)}{\sum_s z_{s,i}+\sum_{u\in \mathcal{U}'} p_{i}(\widetilde{\bm{x}}_u)}.
\end{align}


\subsection{The SALA Algorithm}\label{loss function}

Based on the updated prototype representations $\bm{c}_i$ by the progressive neighbor selection module, we can then compute the prediction  probability  $p_{i}(\bm{x}_q^*)$ for the sample $\bm{x}_q^*$ from the query set $\mathcal{Q}$, where $q=1,2,...,Q$. 

We define the loss function as
\begin{align}\label{loss}
L(\theta,\varphi)= \sum_{q=1}^{Q} \sum_{i=1}^{N} -\mathbb{I}({y}^*_q=i) \log p_{i}(\bm{x}_q^*),
\end{align}
where $y^*_q$ is the label of $\bm{x}_q^*$; $\theta$ and $\varphi$ denote the parameters of the feature embedding module and the SE network in the task-adaptive metric module, respectively. We can train our model jointly in an end-to-end fashion. In Figure \ref{details}, we  show the interaction between the proposed task-adaptive metric module and progressive neighbor selection module.


 We finally summarize SALA in Algorithm \ref{algorithm}.

\begin{algorithm}[ht]
    \caption{SALA (Self-adaptive Label Augmentation for Semi-supervised Few-shot Classification)}\label{alg:SALA}
    \label{algorithm}
    \begin{algorithmic}
        \STATE Randomly initialize $\theta$ and $\varphi$.
        \FOR {$l=1$ \textbf{to} $L$} 
        \STATE Randomly sample one task $\mathcal{S}$, $\mathcal{Q}$ and $\mathcal{U}$.
        \STATE Apply Prototypical Network for $\bm{x}_s \in \mathcal{S}$ to generate the prototype representation $\bm{c_}i$ by Eq. (\ref{cluster}).
        
        \STATE Compute $\bm{\alpha}^T$ with the task-adaptive metric module.
        \FOR {$\widetilde{\bm{x}}_u$ \textbf{in} $\mathcal{U}$}
        \STATE Compute the distance by \eqref{adaptive distance}.
        \ENDFOR
        
        \STATE Apply progressive neighbor selection strategy to update the prototype representations.

        \FOR {$\bm{x}_q^*$ \textbf{in} $\mathcal{Q}$}
        \STATE Compute the prediction result $p_{i}(\bm{x}_q^*)$.
        \ENDFOR
        \STATE Compute the loss function by Eq. (\ref{loss}).
        \STATE Update $\theta$ and $\varphi$ with Adam.
        \ENDFOR
    \end{algorithmic}
\end{algorithm}

\section{Experiments}\label{Experiments}
In this section, we conduct experiments to evaluate the proposed SALA. We first describe the implementation details in Section \ref{implementation details}. Then, we introduce three commonly used datasets for the semi-supervised few-shot classification problem in Section \ref{datasets}. We show the results for semi-supervised few-shot classification with and without distractors in Section \ref{experiment1} and \ref{experiment2}, respectively. Finally, we conduct the ablation study in Section \ref{analysis}.

\subsection{Implementation Details}\label{implementation details}
To compare with the existing methods fairly, we use the four-layer ConvNet as the feature extractor and train SALA from scratch \cite{snell2017prototypical,ren2018meta,liu2019fewTPN}; that is, we do not pre-train SALA on the labeled split of $D_{train}$. We train our model using Adam \cite{kingma2014adam} and the initial learning rate is $10^{-3}$. The largest number of training episodes is set as 500,000, and we evaluate our model on $D_{validation}$ every 2,500 episodes. We will save the parameters of the model once the accuracy reaches the highest in the evaluation dataset.
We perform grid search and select $r=800$ and $\eta=5$ in most cases.

After the training phase, we test our model on $D_{test}$ in similar $N$-way $K$-shot episodes with a larger unlabeled set size to measure the generalization performance of the model following Ren et al. \shortcite{ren2018meta} and Liu et al. \shortcite{liu2019fewTPN}. The reported results are averaged over $1000$ test episodes.

\begin{table*}[ht]
  \caption{Semi-supervised few-shot classification accuracy on three benchmarks without distractor classes.}
  \vskip 0.06in
  \label{miniImageNet}
  \centering
  {\begin{tabular}{lccccc}
    \toprule
  Model & 5-way (Omniglot) &\multicolumn{2}{c}{5-way (\textit{mini}ImageNet)} & \multicolumn{2}{c}{5-way (\textit{tiered}ImageNet)} \\
    & 1-shot&1-shot & 5-shot & 1-shot & 5-shot \\

                    \midrule
                    Soft $k-$Means     & 97.25 & 50.09& 64.59  &51.52&70.25\\
                    Soft $k-$Means+Cluster  & 97.68 & 49.03& 63.08& 51.85&69.42\\                    Masked Soft $k-$Means    & 97.52 & 50.41& 64.39 &52.39&69.88 \\
                    MetaGAN  & 97.58& 50.35 & 64.43 & - &  \multicolumn{1}{c}{-}\\
                    TPN  &-&52.78& 66.42 & 55.74 & 71.01 \\
                \midrule
                    SALA  & \textbf{98.93}& \textbf{54.14}& \textbf{67.23} &\textbf{58.41} &\textbf{72.24}\\
    \bottomrule
  \end{tabular}}
\end{table*}

\subsection{Datasets}\label{datasets}
\textbf{Omniglot}. The Omniglot \cite{lake2011one} dataset consists of 20 instances of 1623 characters from 50 different alphabets. Each instance was drawn by a different person. Following the strategy used in \cite{vinyals2016matching,santoro2016meta,sung2018learning}, we resize the images to 28 $\times$ 28 and use the 1200 classes for training and the remaining 423 classes for testing. In the experiments, we consider 1-shot classification for 5 classes.

\textbf{\textit{mini}ImageNet}. The \textit{mini}ImageNet dataset, originally proposed by \cite{vinyals2016matching}, is a collection of ImageNet \cite{deng2009imagenet} for few-shot classification. The dataset consists of 60,000 colour images with 100 classes, and each class contains 600 examples. We rely on the split of the \textit{mini}ImageNet dataset in Vinvals et al. \shortcite{vinyals2016matching} and Ren et al. \shortcite{ren2018meta}, with 64 classes for training, 16 classes for validation, and 20 classes for test. We consider 1-shot and 5-shot classification for 5 classes in this dataset. All images are resized to 84 $\times$ 84 pixels  following the existing methods \cite{ren2018meta,liu2019fewTPN}.

\textbf{\textit{tiered}ImageNet}. The \textit{tiered}ImageNet dataset is proposed by Ren et al. \shortcite{ren2018meta}. Similar to \textit{mini}ImageNet, this dataset is also a subset of ImageNet. However, it represents a larger subset of ImageNet (608 classes rather than 100 for \textit{mini}ImageNet). Different from \textit{mini}ImageNet, this dataset groups classes into broader categories corresponding to higher-level nodes in ImageNet hierarchy. In this dataset, there are 34 categories totally. We rely on the class split proposed by Ren et al. \shortcite{ren2018meta}, with 20 categories for training (351 classes), 6 categories for validation (97 classes) and 8 categories for test (160 classes). We consider 1-shot and 5-shot classification problems for 5 classes in this dataset and resize the original images to  84 $\times$ 84 pixels \cite{ren2018meta}. 





\subsection{Experiment for Semi-supervised Few-shot Classification without Distractors}\label{experiment1}
In this section, we evaluate the proposed SALA in the standard semi-supervised few-shot classification, where all the unlabeled samples come from the same set of classes as the labeled ones.

\subsubsection{Training Details}
In order to compare with the existing methods directly, we follow the strategies used in previous work \cite{ren2018meta,liu2019fewTPN} to split the images into disjoint labeled and unlabeled sets. For Omniglot and \textit{tiered}ImageNet, we randomly select 10\% of the images of each class as the labeled set, while the remaining 90\% samples as the unlabeled set. For \textit{mini}ImageNet, we sample 40\% of the images of each class as the labeled set and the rest for the unlabeled set. As claimed by Ren et al. \shortcite{ren2018meta}, they choose more data in \textit{mini}ImageNet to form the labeled split because they find that 10\% was too small to avoid over-fitting in the experiments. Moreover, we follow the setup of \textit{episode} \cite{snell2017prototypical,ren2018meta} in our experiments. We first randomly select $N$ classes to form the set of classes. Then, we sample $N\times K$ images in the labeled split to form the support set $\mathcal{S}$, select $Q$ different samples in the labeled split to form the query set $\mathcal{Q}$, and sample $M$ images in the unlabeled split to form the unlabeled set $\mathcal{U}$. 

Following Ren et al. \shortcite{ren2018meta}, for the non-distractor situation, we generate the unlabeled set $\mathcal{U}$ by selecting $H$ samples in the unlabeled split of each $N$ classes in $\mathcal{S}$ ($M=NH$). In the experiments, we adopt $N=5$ following Ren et al. \shortcite{ren2018meta} and apply $H=15$ in most cases.


\subsubsection{Competing Models} In this experiments, we compare the proposed SALA with  several state-of-the-art methods, including three approaches proposed by Ren et al. \shortcite{ren2018meta}, MetaGAN \cite{zhang2018metagan}, and TPN \cite{liu2019fewTPN}.

\subsubsection{Results} 
 In Table \ref{miniImageNet}, we show the performance of our proposed SALA against several state-of-the-art methods in the non-distractor case. From the results, we can see that SALA significantly outperforms the state-of-the-arts under all settings.
 Specifically, for 5-way 1-shot tasks on \textit{mini}ImageNet and \textit{tiered}ImageNet, the improvement of SALA over the state-of-the-arts reported by Ren et al. \shortcite{ren2018meta}---which are all metric-based methods---is up to 3.73\% and 6.02\%, which demonstrates the effectiveness of our proposed method. Moreover, we find that the absolute improvement of accuracy in 1-shot tasks is more significant than that in 5-shot tasks. A possible reason is that the prototypical presentations would be more discriminative with more labeled instances, which demonstrates that SALA is more effective in the extremely few-shot situation (e.g., 1-shot classification).

\subsection{Experiment for Semi-supervised Few-shot Classification with Distractors}\label{experiment2}
In this section, we evaluate the proposed SALA in a more challenging case, where the unlabeled samples in $\mathcal{U}$ may come from classes unseen in $\mathcal{S}$.
 
\subsubsection{Training Details} 
In this experiment, we follow the split and selection strategy described in Section \ref{experiment1}. Moreover, when including distractor classes, we additionally sample $C$ other classes from the dataset and $H$ samples from the unlabeled split of each to form the unlabeled set $\mathcal{U}$; that is $M=NH+CH$. In the experiments, we use $C=N=5$ following Ren et al. \shortcite{ren2018meta} and apply $H=15$ in most cases. When we apply the progressive neighbor selection module, we set $M_0=M/2$ in the experiment. 

\subsubsection{Competing Models} Notice that, MetaGan \cite{zhang2018metagan} and TPN \cite{liu2019fewTPN} do not consider this more challenging case where distractors are available. Therefore, we compare SALA with three methods proposed by Ren et al. \shortcite{ren2018meta} in this experiment.

\begin{table*}[ht]
  \caption{Semi-supervised few-shot classification accuracy on three benchmarks with distractor classes.}
    \vskip 0.06in
  \label{distractor}
  \centering
  {\begin{tabular}{lccccc}
    \toprule
   Model & 5-way (Omniglot) &\multicolumn{2}{c}{5-way (\textit{mini}ImageNet)} & \multicolumn{2}{c}{5-way (\textit{tiered}ImageNet)} \\
    & 1-shot&1-shot & 5-shot & 1-shot & 5-shot \\
    \midrule
                    Soft $k-$Means     & 95.01 & 48.70 & 63.55  &49.88&68.32  \\
                    Soft $k-$Means+Cluster & 97.17 & 48.86 & 61.27& 51.36 &67.56 \\
                    Masked Soft $k-$Means   & 97.30  & 49.04 & 62.96 &51.38&69.08 \\
                \midrule
                    SALA  & \textbf{98.16 }& \textbf{51.44}& \textbf{64.12} &\textbf{55.85} &\textbf{69.45}\\
    \bottomrule
  \end{tabular}}
\end{table*}

\subsubsection{Results} 
We show the performance of SALA against the competing methods in Table \ref{distractor}.
SALA reaches the accuracy of 98.16\% on Omniglot, 51.44\% and 64.12\% on \textit{mini}ImageNet, and 55.85\% and 69.45\% on \textit{tiered}ImageNet. We can see that SALA significantly outperforms the state-of-the-art methods, which demonstrates that our proposed approach can also effectively address this more challenging case. Specifically, SALA surpasses the second best model with an absolute margin of 4.47\% on \textit{tiered}ImageNet in 1-shot tasks. Moreover, the absolute improvement of SALA in 1-shot tasks is still more significant than that in 5-shot tasks, which is consistent with the results in Table \ref{miniImageNet}.

\subsection{Ablation Study and Analysis}\label{analysis}
In this section, we first conduct ablation studies of SALA. Then, we perform the analysis of the proposed two modules.

\subsubsection{Ablation Study}
We study the impact of the task-adaptive metric module and the progressive neighbour selection module on \textit{mini}ImageNet and \textit{tiered}ImageNet under the 5-way 1-shot setting. The results are shown in Table \ref{ablation}.

From the results in Tabel \ref{ablation}, we can see that both the proposed modules can effectively improve the results of the baseline. Moreover, we can find that the major gain of the improvement comes from the task-adaptive metric module, and we can obtain an additional gain by using the progressive neighbor selection strategy.


\begin{figure}[ht]
    \centering{
                \subfigure[]{ \label{class_aware1}
            \includegraphics[height=0.35\columnwidth]{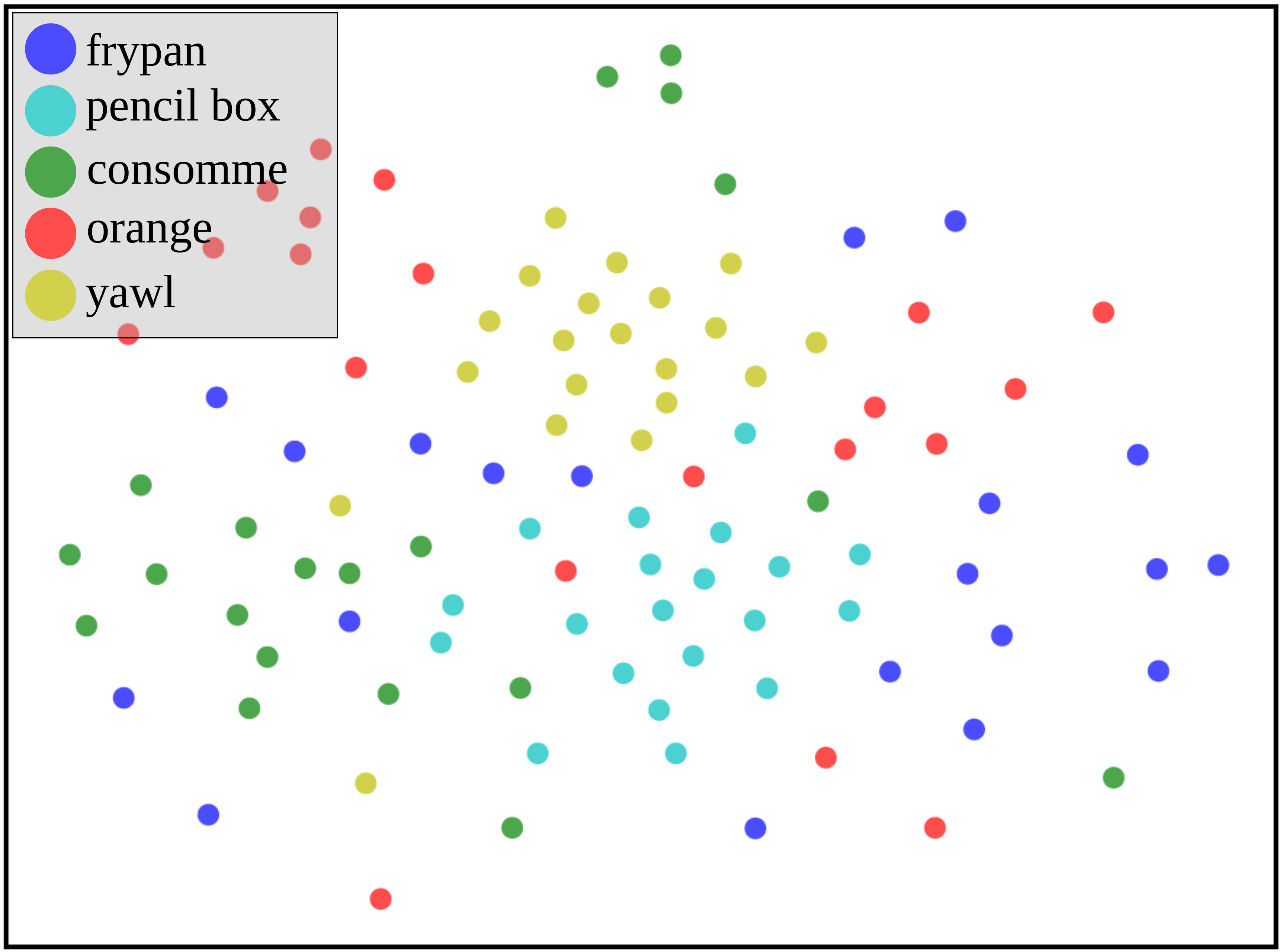}
        }
        \subfigure[]{ \label{class_aware2}
            \includegraphics[height=0.35\columnwidth]{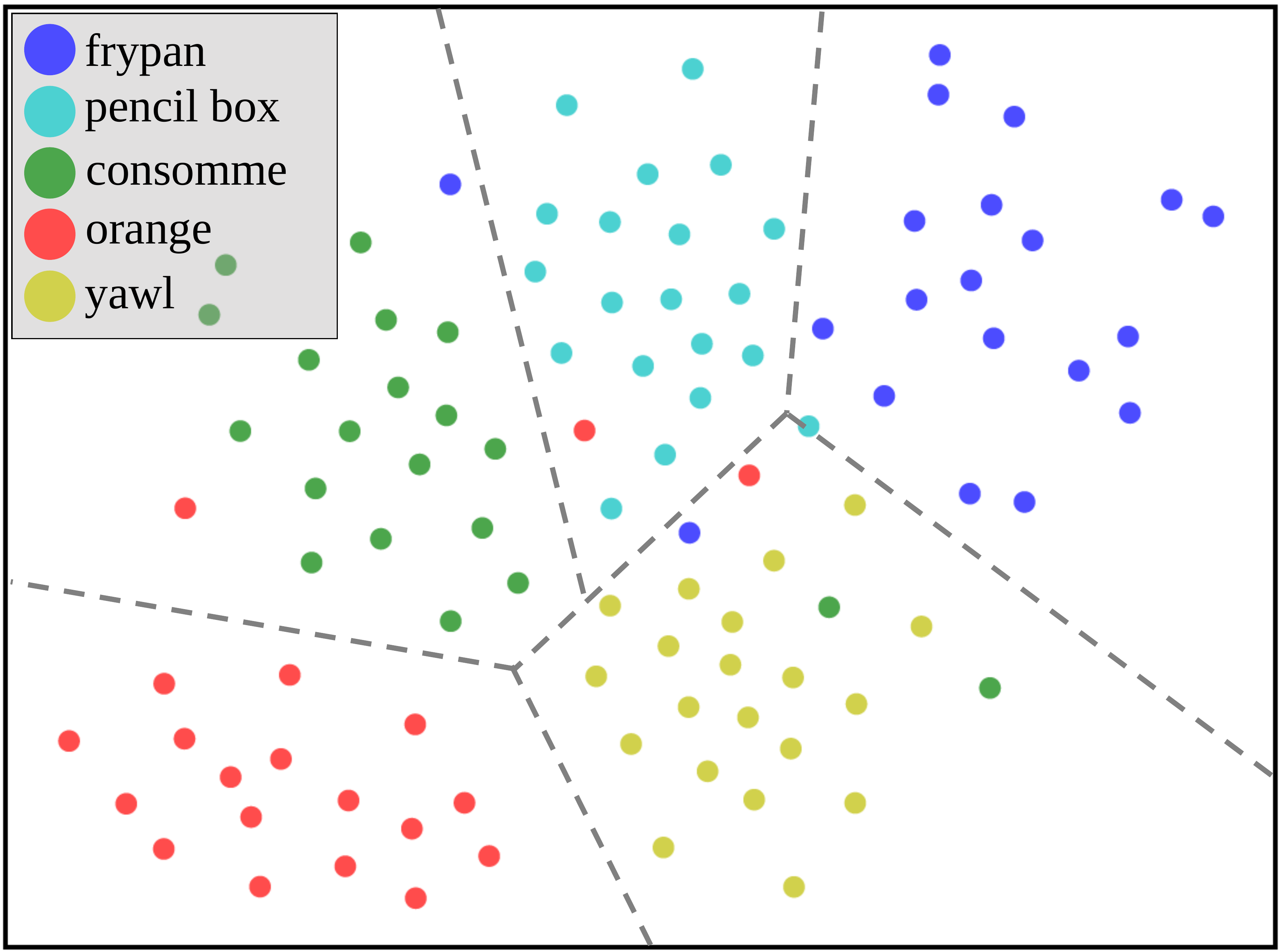}
        }
    }
    \caption{An example of the embedding visualization before and after the proposed task-adaptive metric module. (a) Original embeddings. (b) Embeddings after the self-adaptive metric module.}
    \vskip -0.1in
    \label{visual}
\end{figure}

\begin{table}[]
\vskip -0.1in
  \caption{Ablation study of SALA. We report the classification accuracy on \textit{mini}ImageNet and \textit{tiered}ImageNet under the 5-way 1-shot setting. TM: task-adaptive metric module. PNS: progressive neighbor selection module.}
  \vskip 0.06in
  \label{ablation}
  \centering
  \begin{tabular}{p{0.8cm}<{\centering}p{0.8cm}<{\centering}p{2.5cm}<{\centering}p{2.5cm}<{\centering}}
    \toprule
    TM & PNS &\textit{mini}ImageNet & \textit{tiered}ImageNet \\
    \midrule
&&51.73 $\pm$ 0.72&55.76 $\pm$ 0.89\\
                       & \checkmark& 52.23 $\pm$ 0.90 & 57.08 $\pm$ 0.93\\
                    \checkmark &&53.70 $\pm$ 0.73 & 58.04 $\pm$ 0.97\\
           \midrule            
                   \checkmark &\checkmark & \textbf{54.14 $\pm$ 0.87} & \textbf{58.41 $\pm$ 0.92}\\
    \bottomrule
  \end{tabular}
  \vskip -0.08in
\end{table}
\begin{table}[!h]
\vskip -0.08in
  \caption{Accuracy of TADAM and SALA on \textit{mini}ImageNet and \textit{tiered}ImageNet in the semi-supervised few-shot classification without distractor classes.}
  \vskip 0.06in
  \label{tadamcompare}
  \centering
  \begin{tabular}{p{2cm}p{2.5cm}<{\centering}p{2.5cm}<{\centering}}
    \toprule
    Model &\textit{mini}ImageNet & \textit{tiered}ImageNet \\
    \midrule
                    TADAM& 51.03 $\pm$ 0.83 & 55.21 $\pm$ 0.62\\
                   SALA& \textbf{53.70 $\pm$ 0.73} & \textbf{58.04 $\pm$ 0.97} \\
    \bottomrule
  \end{tabular}
  \vskip -0.08in
\end{table}

\subsubsection{Analysis of Task-adaptive Metric Module} 
To illustrate that the proposed task-adaptive metric module can effectively encodes the intrinsic data property for different classes, we show the embedding visualization (projected to a 2D plane by t-SNE \cite{maaten2008visualizing}) before and after this module in \figref{visual}. \figref{class_aware1} shows the original embeddings. \figref{class_aware2} shows the embeddings which multiply $\bm{\alpha}^T$ with the original features. From \figref{visual}, we can see that the boundaries between different classes are more clear after our module, which implies that our learned task-adaptive module can effectively capture the most discriminative features for different classes and different tasks.

Moreover, we compare our task-adaptive metric with TADAM \cite{oreshkin2018tadam}, which propose the metric scaling strategy to improve the performance. We conduct experiments on two benchmarks in the semi-supervised 5-way 1-shot classification without distractors. The results are shown in Table \ref{tadamcompare}. From the results, we can see that our proposed task-adaptive metric module significantly outperforms TADAM.




\begin{figure}[]
    \centering{
                \subfigure[\textit{mini}ImageNet]{ \label{1}
            \includegraphics[height=0.401\columnwidth]{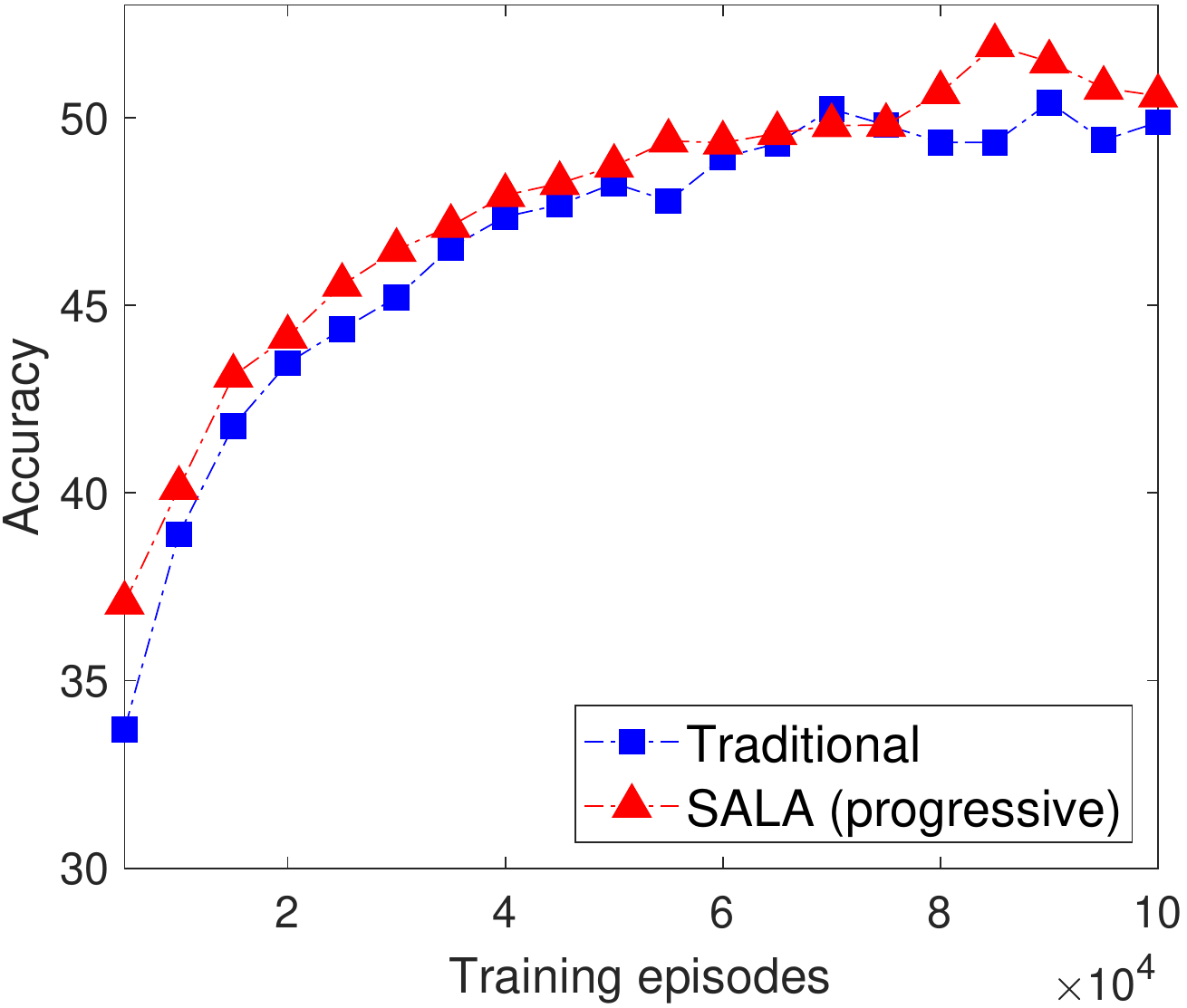}
        }
        \subfigure[\textit{tiered}ImageNet]{ \label{2}
            \includegraphics[height=0.401\columnwidth]{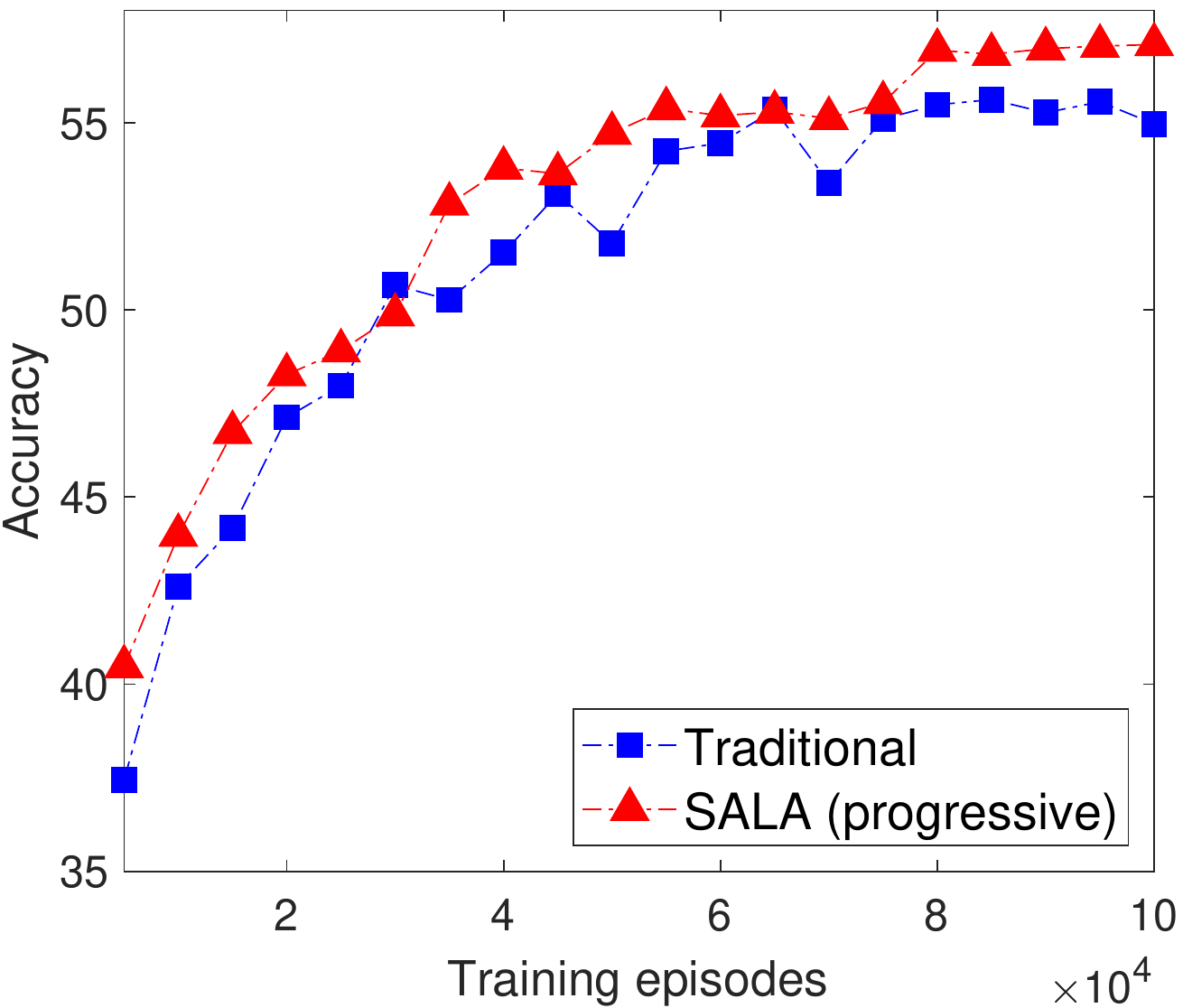}
        }
    }
    \caption{Accuracy of the progressive neighbor selection module and traditional sampling strategy on \textit{mini}ImageNet and \textit{tiered}ImageNet for the first 100,000 training episodes.}
    \vskip -0.1in
    \label{fig:time}
\end{figure}

\subsubsection{Analysis of Progressive Neighbor Selection Module} We show the accuracy of the progressive neighbor selection module and the traditional strategy for the first 100,000 training episodes in \figref{fig:time}. The traditional strategy uses all unlabeled data in each episode to update the prototype representations. We conduct experiments on \textit{mini}ImageNet and \textit{tiered}ImageNet datasets under 5-way 1-shot setting. As the training proceeds, we can see that our approach outperforms the traditional sampling strategy consistently, which demonstrates the effectiveness of the proposed module. Moreover, we find that the absolute improvement of accuracy on \textit{tiered}ImageNet is larger than that on \textit{mini}ImageNet, which is consistent with the results in Tabel \ref{ablation}.



\section{Conclusion}\label{Conclusion}

In this paper, we propose a self-adaptive label augmentation algorithm, called SALA, for semi-supervised few-shot classification. The major novelty of SALA is the proposed task-adaptive metric that can capture the intrinsic property of different tasks. Moreover, inspired by the observation that the pseudo-labels will be more reliable as the training course proceeds, we introduce a novel progressive neighbor selection strategy to select unlabeled data with the most reliable pseudo labels for training. Experiments show that the proposed SALA achieves the state-of-the-art performance. One of our future directions is to extend our SALA to other important real applications, such as machine translation, dialogue system, knowledge graph, and query understanding.

\bibliographystyle{aaai}
\bibliography{paper}

\end{document}